\documentclass[sigconf]{acmart}
\AtBeginDocument{%
  \providecommand\BibTeX{{%
    \normalfont B\kern-0.5em{\scshape i\kern-0.25em b}\kern-0.8em\TeX}}}

\copyrightyear{2022}
\acmYear{2022}
\setcopyright{rightsretained}
\acmConference[WWW '22 Companion]{Companion Proceedings of the Web Conference 2022}{April 25--29, 2022}{Virtual Event, Lyon, France}
\acmBooktitle{Companion Proceedings of the Web Conference 2022 (WWW '22 Companion), April 25--29, 2022, Virtual Event, Lyon, France}\acmDOI{10.1145/3487553.3524234}
\acmISBN{978-1-4503-9130-6/22/04}

\usepackage{graphicx}
\usepackage{diagbox}
\begin{document}

\title{Cross-Language Learning for Product Matching}

\author{Ralph Peeters}
\orcid{0000-0003-3174-2616}
\affiliation{%
  \institution{Data and Web Science Group}
  \institution{University of Mannheim}
  \city{Mannheim}
  \country{Germany}
}
\email{ralph@informatik.uni-mannheim.de}

\author{Christian Bizer}
\orcid{0000-0003-2367-0237}
\affiliation{%
  \institution{Data and Web Science Group}
  \institution{University of Mannheim}
  \city{Mannheim}
  \country{Germany}
}
\email{chris@informatik.uni-mannheim.de}

\renewcommand{\shortauthors}{Peeters and Bizer}

\begin{abstract}
Transformer-based entity matching methods have significantly moved the state of the art for less-structured matching tasks such as matching product offers in e-commerce. In order to excel at these tasks, Transformer-based matching methods require a decent amount of training pairs. Providing enough training data can be challenging, especially if a matcher for non-English product descriptions should be learned. This poster explores along the use case of matching product offers from different e-shops to which extent it is possible to improve the performance of Transformer-based matchers by complementing a small set of training pairs in the target language, German in our case, with a larger set of English-language training pairs. Our experiments using different Transformers show that extending the German set with English pairs improves the matching performance in all cases. The impact of adding the English pairs is especially high in low-resource settings in which only a rather small number of non-English pairs is available. As it is often possible to automatically gather English training pairs from the Web by exploiting schema.org annotations, our results are relevant for many product matching scenarios targeting low-resource languages. 
\end{abstract}



\maketitle

\section{Introduction}

Identifying offers for the same product is one of the central challenges in e-commerce applications such as price comparison portals and electronic marketplaces. Training Transformer-based matchers using offers from different e-shops which share the same product identifier has proven to be a successful solution for product matching reaching F1 scores above 0.9 in many cases~\cite{peeters2021,primpeli_wdc_2019,liDeepEntityMatching2020}. The bottleneck of this approach is that it requires a decent amount of pairs of offers for the products to be matched as training data. In recent years, large numbers of websites have started to markup structured data within their pages using the schema.org vocabulary\footnote{\url{https://schema.org/}}. One of the most widely annotated entity types are product offers: Analyzing the October 2021 CommonCrawl web corpus, the WebDataCommons project for instance found 1.5 million websites to annotate product offers within their pages\footnote{\url{http://webdatacommons.org/structureddata/\#toc3}}. This means that for widely-used languages such as English, the required training data can be extracted from web crawls by relying on schema.org annotations which identify product titles, product descriptions, and product identifiers such as GTIN or MPN numbers within web pages\footnote{\url{http://webdatacommons.org/largescaleproductcorpus/v2/}}~\cite{primpeli_wdc_2019}. For less widely used languages and less commonly sold products, it can be difficult to find enough offers in the respective target language on the Web. 

To deal with this problem, this poster explores the utility of English language offers for training product matchers for less widely spoken target languages, such as German. For this, we experiment with training sets combining a large amount of English language offer pairs with smaller amounts of training pairs in the target language. We experiment with matchers that internally rely on different pre-trained Transformer models including BERT~\cite{devlinBERTPretrainingDeep2019}, a German version of BERT\footnote{https://github.com/dbmdz/berts}, multilingual BERT\footnote{https://github.com/google-research/bert/blob/master/multilingual.md},  XLM-R~\cite{conneau2019unsupervised}, as well as on an SVM classifier. The multi-lingual Transformers were originally pre-trained on multi-lingual text and are fine-tuned without using explicit cross-language alignments from multi-lingual dictionaries~\cite{ruderSurveyCrosslingualWord2019,wangCrosslingualKnowledgeGraph2018}. Recent work on aligning cross-language embeddings for entity linking can be found in the area of knowledge graph embeddings~\cite{wangCrosslingualKnowledgeGraph2018,chenCotrainingEmbeddingsKnowledge2018,peiREARobustCrosslingual2020}.

Our experiments show that extending the German training set with English pairs is always beneficial. The impact of adding the English pairs is especially high in low-resource settings in which only a small set of German pairs is available. 
The contributions of the poster are twofold: 1. Up to our knowledge, we are the first to experiment with cross-language learning for matching textual entity descriptions using Transformers. 2. We demonstrate that by combining English language training pairs and a rather small amount of training pairs in the target language, it is possible to learn product matchers reaching F1 scores above 90\%, clearly outperforming matchers which were  trained using only product offers in the target language.


\begin{figure*}[htb]
  \centering
  \includegraphics[width=0.8\linewidth]{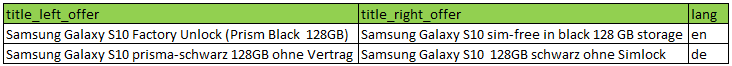}
  \caption{Example of a matching product offer pair in English (top) and a pair for the same product in German (bottom).}
  \Description{Example of a matching product offer pair in English language (top) and a pair for the same product in German (bottom).}
  \label{fig:pair-example}
\end{figure*}

\section{Datasets}
\label{sec-datasets}

We experiment with English- and German-language offers for mobile phones which have been crawled from 66 different e-shops, auction platforms, and electronic marketplaces. Each offer contains a title, a description, and some product identifier such as a GTIN or MPN number. The data was collected using 150 mobile phones as seeds for the crawling process. These seeds contained widely-sold head products but also less-sold long-tail phones. In addition to offers for the seed phones, the dataset also contains offers for further phones which have been discovered during the crawling process. We group the offers into pairs by using shared GTIN, EAN, and MPN numbers as distant supervision~\cite{primpeli_wdc_2019}.

\begin{figure}[htb]
  \centering
  \includegraphics[width=\linewidth]{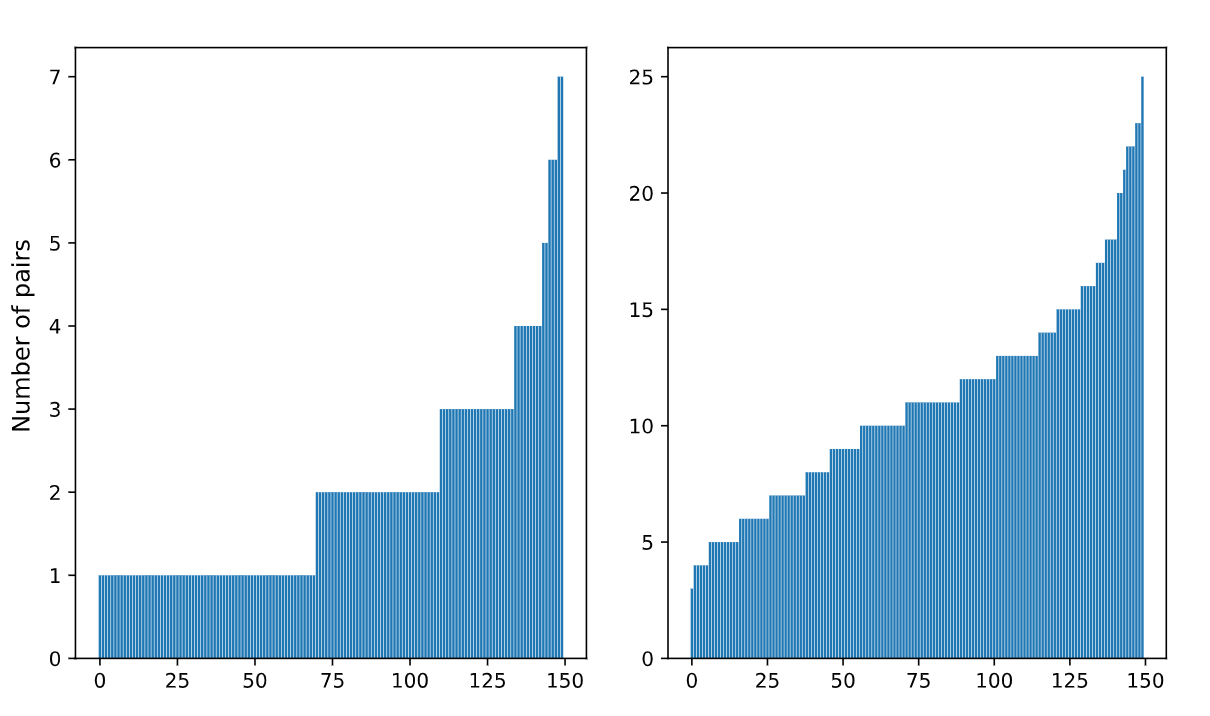}
  \caption{Distribution of positive (left) and negative (right) pairs in the German test set for each of the 150 seed mobile phones.}
  \Description{Distribution of positive (left) and negative (right) pairs in the German test set for each of the 150 seed mobile phones.}
  \label{fig:distribution}
\end{figure}

Afterward, we remove the identifiers from the offers in order to prevent the matching from being trivial. Non-matching pairs were created by combining an offer for a seed product with an offer for a similar seed product or a similar offer from a phone which has been discovered during crawling.  We arrange the pairs into language-specific training sets of different sizes. The training sets range from 450 to 7200 pairs and contain 50\% matches and 50\% non-matches. Figure \ref{fig:pair-example} shows an example of a pair of English product offers and a pair for the same product in German. We use a German-language test set containing 1200 pairs (25\% matches, 75\% non-matches). None of the pairs in the test set is also contained in the training sets. Half of the pairs in all sets were chosen randomly, while the other half contains corner-case pairs (measured by cosine similarity). Figure \ref{fig:distribution} shows the distribution of positive and negative pairs in the German test set for the 150 seed mobile phones.
This webpage\footnote{http://data.dws.informatik.uni-mannheim.de/Web-based-Systems-Group/StudentProjects/2020/Cross-lingual-Product-Matching-using-Transformers/} provides additional information about the dataset creation process, offers for the same products in additional languages (Spanish and French), as well as the code for replicating the experiments.

\section{Models and Baselines}
\label{sec-models}


We utilize three different pre-trained Transformer-based models from the HuggingFace\footnote{https://huggingface.co/transformers/} library. As a monolingual English option, we use the BERT base model ('bert-base-uncased'). The model was pre-trained on the English Wikipedia and BookCorpus. We further use a German BERT model ('bert-base-german-dbmdz-uncased') pre-trained on a diverse set of German texts including the German Wikipedia, parts of the Common Crawl, and the EU Bookshop corpus, which roughly sums up to the same amount of pre-training data as used for the English BERT base model. As multilingual models, we use multi-lingual BERT ('bert-base-multilingual-uncased'), which is trained on the top 100 largest Wikipedias, as well as XLM-RoBERTa ('xlm-roberta-base'), which is trained on a CommonCrawl corpus consisting of 100 different languages. This selection of models allows us to examine the performance gain resulting from pre-training using multilingual texts. Furthermore, we calculate a simple word co-occurrence baseline using an SVM classifier for predictions.

We create the input sequences for the Transformer-based models by concatenating the title and description attributes of a product offer into one string and applying the respective tokenizer to both product offers in a pair to represent them in the standard input representation for Sequence Classification, i.e. "[CLS] Product 1 [SEP] Product 2 [SEP]" for BERT-based models. As input for the SVM baseline, we generate a bag-of-words word vector representation indicating co-occurring words in a product pair, which serve as input for the classifier. 

For every experimental run, the learning rate was optimized in the range between 5e-6 and 1e-4 using a validation set and early stopping. If a given model did not improve for three consecutive epochs during hyperparameter tuning, the run was stopped. During training, we fine-tuned the models for 25 epochs. We utilized a fixed batch size of 16 and a weight decay of 0.01. All other hyperparameters were set to their default values. The scores reported are averages over three runs that were individually trained using the same hyper-parameter setup.


\begin{table}[htb]
\centering
\caption{Results on the German test set w/ and w/o additional English training data. Training sizes: EN - 7200, DE - 1800.}
\label{tab:method-comparison-static}
\resizebox{0.47\textwidth}{!}{%
\begin{tabular}{|l|c|c|c|c|c|}
\hline
              & SVM   & BERT  & gBERT & mBERT          & XLM-R \\ \hline
F1 without EN & 71.00 & 65.27 & 73.43 & 87.69          & 73.40     \\ \hline
F1 with EN    & 71.05 & 74.29 & 89.83 & \textbf{91.44} & 86.98     \\ \hline
Difference    & 0.05  & 9.02  & 16.40 & 3.75           & 13.58     \\ \hline
\end{tabular}%
}
\end{table}

\section{Results and Discussion}
\label{sec-results}

In the first set of experiments, we compare the performance of the different mono- and multi-lingual models on the German test set while training on the one hand with only 1800 German training pairs and on the other hand with the same 1800 German pairs and an additional 7200 English pairs. Table \ref{tab:method-comparison-static} shows the results of this experiment. When fine-tuned with only 1800 German pairs, the English BERT model scores the lowest overall, falling 6\% F1 behind the SVM baseline at 65\% F1, meaning that the language misfit between English pre-training and German fine-tuning has a severe negative impact. The German version of BERT can improve on the SVM by 2.5\% F1, showing the importance of German language pre-training when compared to English BERT. The multilingual XLM-R achieves a comparable result to German BERT. Multilingual BERT achieves the best result with 87\% F1, outperforming German BERT by 14\% F1. The very strong performance of multilingual BERT in this case likely stems from the larger amount of pre-training data the model was trained on (Wikipedia dumps in 100 languages) in comparison to most other models. Even though only a part of its training data is in German, the large and varied pre-training data across different languages leads to a model that is highly training data efficient when being fine-tuned in German. After extending the German training set with the additional 7200 English training pairs (Row \textit{F1 with EN} in Table \ref{tab:method-comparison-static}), all models apart from the SVM improved significantly. German BERT sees the highest improvement with a gain of 16\% F1 putting it 1.5\% points behind multilingual BERT, which achieves the overall highest score of 91.4\% F1. These results clearly show that adding training data in a high-resource language like English to a smaller amount of training data in the target language leads to improvements for all transformer models and thus is a promising course of action for practitioners. 

\begin{table}[htb]
\centering
\caption{Results on the German test set when varying the amount of training data for both languages.}
\label{tab:change-sizes}
\resizebox{0.47\textwidth}{!}{%
\begin{tabular}{|c|c|c|c|c|c|c|c|}
\hline
\diagbox{DE}{EN} & 0     & 450   & 900   & 1800  & 3600  & 7200  & $\Delta$ 0-7200 \\ \hline
450  & 67.11 & 72.79 & 75.44 & 80.83 & 86.82 & 87.97 & 20.86         \\ \hline
900  & 75.76 & 75.10 & 74.00 & 87.67 & 88.92 & 88.19 & 12.43         \\ \hline
1800 & 87.69 & 88.43 & 88.38 & 90.17 & 90.72 & 91.44 & 3.75          \\ \hline
3600 & 93.63 & 92.98 & 92.46 & 93.97 & 93.25 & 94.46 & 0.83          \\ \hline
\end{tabular}%
}
\end{table}

In a second set of experiments, we train mBERT with different combinations of training data sizes for both languages in an effort to understand how much training data in the target language is required and how much English training data needs to be added in order to reach a high performance level. Table \ref{tab:change-sizes} shows the results of the experiments. For German training sets of size 900 and less, any combination of training data sets resulting in an overall amount of less than 2000 pairs leads to an F1 around 75\% F1 or less regardless of the composition among languages. If the training set consists of more than 2000 pairs, an F1 over 80\% is achievable in all scenarios. If training data in the target language is scarce (450 pairs) adding English training pairs has a significant effect, leading to a large improvement in every step up until 3600 additional English pairs (86.82\% F1). After this point, doubling the amount of English training data only yields an improvement of 1\% F1. The beneficial effect of additional English training data is visible across all settings though it diminishes with increasing size of the German training set. Once the German training set reaches a size of 1800 pairs, the effect of adding English training data is no longer as strong as before but still results in an overall improved model, reaching a maximum of 94.5\% F1 when trained using 3600 German and 7200 English pairs.

\section{Conclusion}
\label{sec-conclusion}

We have shown that the performance of Transformer-based product matchers for low-resource languages can be significantly improved by adding English-language offer pairs to the training set. The impact of adding the English pairs is especially high for low-resource settings in which only a rather small number of non-English pairs is available. It further turned out that for successful cross-language learning during fine-tuning, a Transformer model that has also been pre-trained on large amounts of text in different languages should be chosen as starting point. Given that training pairs in head-languages such as English can automatically be extracted from the Web by exploiting schema.org annotations~\cite{primpeli_wdc_2019}, we believe that cross-language learning can contribute to reducing labeling costs in many low-resource language matching scenarios. 

\begin{acks}
The datasets used in this poster were assembled by a student team consisting of Andreas K\"upfer, Benedikt Ebing, Daniel Schweimer, Fritz Niesel, and Jakob Gutmann.
\end{acks}

\bibliographystyle{ACM-Reference-Format}
\bibliography{main}

\end{document}